%% file: bare_conf.tex
\documentclass[a4paper,conference]{IEEEtran}
\ifCLASSINFOpdf
\else
\fi

\usepackage{graphicx}
\usepackage{subcaption}
\usepackage{tikz, pgfplots}
\usepackage{algorithm}
\usepackage{algorithmic}
\usepackage{amsfonts}
\usepackage{amssymb}
\usepackage{amsmath}
\usepackage[normalem]{ulem}
\usepackage{color}
\usepackage{xspace}
\begin{document}
%
% paper title
% Titles are generally capitalized except for words such as a, an, and, as,
% at, but, by, for, in, nor, of, on, or, the, to and up, which are usually
% not capitalized unless they are the first or last word of the title.
% Linebreaks \\ can be used within to get better formatting as desired.
% Do not put math or special symbols in the title.
\title{Tackling Occlusion in Siamese Tracking \\ with Structured Dropouts}

% author names and affiliations
% use a multiple column layout for up to three different
% affiliations
\author{\IEEEauthorblockN{Deepak K. Gupta, Efstratios Gavves and Arnold W. M. Smeulders}
\IEEEauthorblockA{QUVA Lab \& Informatics Institute, University of Amsterdam\\
Amsterdam, The Netherlands\\
Email: \{D.K.Gupta, E.Gavves, A.W.M.Smeulders\}@uva.nl}}

% make the title area
\maketitle

% As a general rule, do not put math, special symbols or citations
% in the abstract
\begin{abstract}
\input{abstract}
\end{abstract}

% no keywords

% For peer review papers, you can put extra information on the cover
% page as needed:
% \ifCLASSOPTIONpeerreview
% \begin{center} \bfseries EDICS Category: 3-BBND \end{center}
% \fi
%
% For peerreview papers, this IEEEtran command inserts a page break and
% creates the second title. It will be ignored for other modes.
\IEEEpeerreviewmaketitle

\section{Introduction}
\input{intro}
% no \IEEEPARstart

\section{Background}
\input{background}

\section{Structured Dropouts for Modeling Occlusion}
%\subsection{Proposed Dropouts}
\input{propose_dropouts}

%\subsection{Explicit and Implicit Inference}
%\input{inference}

\section{Related Work}
\input{related_work}

\section{Experiments}
We demonstrate the efficacy of the structured dropouts through a series of experiments on several tracking benchmarks. 

\subsection{Baselines} 
\input{tracker_choice}

\subsection{Data}
\input{data}

\subsection{Implementation details}
\label{sec_imp_details}
\input{imp_details}

\subsection{Results}
\input{results}

\subsection{Discussion}
\input{discuss}

\section{Conclusions}
\input{conclusion}

\bibliographystyle{IEEEtran}
% argument is your BibTeX string definitions and bibliography database(s)
\bibliography{egbib}
%
% <OR> manually copy in the resultant .bbl file
% set second argument of \begin to the number of references
% (used to reserve space for the reference number labels box)

% that's all folks
\end{document}

%% file: abstract.tex
Occlusion is one of the most difficult challenges in object tracking to model. This is because unlike other challenges, where data augmentation can be of help, occlusion is hard to simulate as the occluding object can be anything in any shape. 

In this paper, we propose a simple solution to simulate the effects of occlusion in the latent space. Specifically, we present structured dropout to mimick the change in latent codes under occlusion. We present three forms of dropout (channel dropout, segment dropout and slice dropout) with the various forms of occlusion in mind. To demonstrate its effectiveness, the dropouts are incorporated into two modern Siamese trackers (SiamFC and SiamRPN++). The outputs from multiple dropouts are combined using an encoder network to obtain the final prediction. Experiments on several tracking benchmarks show the benefits of structured dropouts, while due to their simplicity requiring only small changes to the existing tracker models.

%% file: intro.tex
\begin{figure}
\centering
\begin{subfigure}{0.23\textwidth}
    \centering 
    \includegraphics[height=2cm, width=2.7cm]{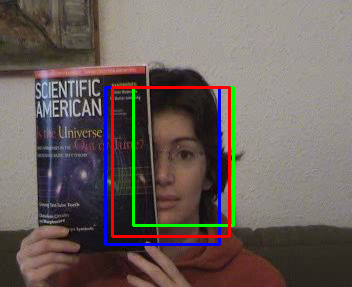}
\end{subfigure}
\begin{subfigure}{0.23\textwidth}
    \centering 
    \includegraphics[height=2cm, width=2.7cm]{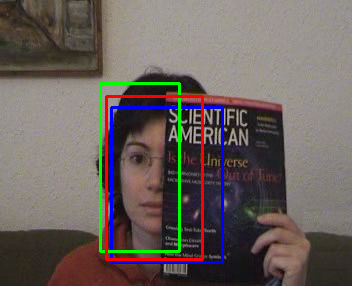}
\end{subfigure}
\begin{subfigure}{0.23\textwidth}
    \centering 
    \includegraphics[height=2cm, width=2.7cm]{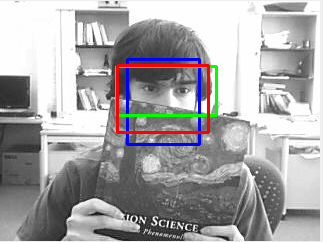}
\end{subfigure}
\begin{subfigure}{0.23\textwidth}
    \centering 
    \includegraphics[height=2cm, width=2.7cm]{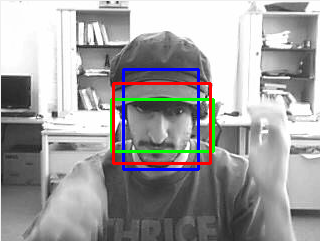}
\end{subfigure}
\caption{Example image frames from OTB100 dataset \protect\cite{Wu2015cvpr} showing the ground-truth bounding boxes (blue), and the predictions obtained using SiamRPN++\protect\footnotemark \protect\cite{Li2019cvpr} (green) and our SiamRPN++ with structured dropouts (red). }
\label{fig_schem1}
\end{figure}

% ARN: Agreed with Stratis. Edited to be much more direct. Check for technical correctness.
% \strr{Verbose. Better go directly to the point of the paper, that is occlusion. Everyone in a vision and learning conferences (should) know what object tracking is. Focus on what is your problem and what new do you bring to the table.}{Object tracking refers to the problem of identifying the state of an object of interest in each frame of a given video based on an accurate initial state provided in the first frame.} 
Recent trackers, just to name a few, SINT \cite{Tao2016cvpr}, SiamFC \cite{Bertinetto2016eccvw}, SiamRPN \cite{Li2018cvpr}, SiamRPN++ \cite{Li2019cvpr} and DiMP \cite{Bhat2019iccv} profit from  learning to discriminate under varying conditions of illumination, change in shape or size or clutter. The starting point of this paper is that  learning more does not help to discriminate in occlusion, see Figure \ref{fig_schem1}, occlusion has to be learned in a better way.
%\str{I think the 1st par is irrelevant to the problem statement. Do we discuss the limitations of deep architectures in tracking? I think it is better to delete the 1st par and use the 2nd par as a 1st par.} \dpk{I am keeping this paragraph, since it says that the direction in which most of the development is not directly aligned with handling occlusion, and it refers to an example to start the story.} 
% AS: Do not be stubborn, it serves no one. As a general rule, if an intelligent person takes the time to make a comment, then do something: either improve the argumentation or change it.

Most recent trackers are based on the concept of Siamese matching, starting from \cite{Tao2016cvpr}, \cite{Bertinetto2016eccvw}, and evolving into \cite{Li2019cvpr} as a good example of the state of the art. However good the tracker is, commonly fails to handle  occlusion. This is attributed to the fact that confidence scores of current Siamese trackers match the complete image of the target to the query. When doing so, the occluded parts of the target will degrade the confidence score of the prediction. 

%{\color{green} ARN can we say from studying their performance that occlusion is the most often cause of error of SiamRPN++? This would make the paper much more valuable.}\dpk{Occlusion is not the main cause, hence, I would keep it in a soft tone that occlusion is among the challenges.}

%While there are several other developments that have contributed towards improving the performance of trackers, the choice of the backbone network has played a crucial role in governing the performance of a deep learning-based tracker. In a recent work \cite{}, it has been demonstrated that using  the Resnet-101 architecture in the backbone can lead to further improvement of performance. Clearly, one of the ways to improve the discriminative power of trackers is to use a better backbone model, and better mostly implies a deeper model.

\footnotetext{The implementation of SiamRPN++ has been obtained from https://github.com/STVIR/pysot.}

A straightforward and common approach to tackle appearance variations is to augment the training set of the target-to-query similarity function with variations. For illumination variations, one uses darker and lighter variations of the same image, and the same holds for shifted \cite{Tao2016cvpr}, rotated \cite{Worrall2017cvpr} and scaled \cite{Sosnovik2020iclr} versions of the target. The effect is that the similarity function is less variant to accidental effects. However, it is fundamentally hard to simulate realistic occlusions as the nature of the occlusion is different each time. The occlusion group is infinitely big in its appearance as the occluded object can be anything. Therefore, an approach based on a large dataset with explicit annotations of occlusions is likely to fail as no one dataset could cover all appearance variations caused by occlusions. The only common denominator is the ever-decreasing visible target area, and that observation is what we take as our starting point.

In this paper, we tackle occlusion through exploiting the inherent structure of Siamese networks. We propose structured dropouts, inspired by \cite{Gal2016icml} who use ensembles of dropout models for Bayesian predictions. Rather than applying dropouts using independent Bernoulli distributions per feature map locations, we sample them from spatially or channel-wise structured probabilities. The outputs from multiple passes of structured dropout are combined using an encoder architecture to obtain the final prediction. Experiments on Siamese trackers equipped with structured dropouts demonstrate improved performance on several tracking benchmarks.
%Note that we do not claim that our structured dropouts are absolute representations of occlusion. 
%\ssout{These formulations are based on the interpretation of encoding framework in the twin subnetworks of Siamese trackers. Different variants of the structured dropouts are applied along spatial as well as channel dimensions of the convolutional networks with interpretations related to different forms of occlusion. }

%\strr{I find the following paragraph redundant. The explicit and implicit mode are not exclusive. You can apply both of them. In that sense, I would delete the par so that it does not look like we are overselling. Better describe this simply as an extra paragraph by the end of the model section.}
%{We also propose two different approaches (\emph{explicit} and \emph{implicit}) to perform inference over results from multiple passes of proposed dropout sampling methods. In explicit inference, structure dropouts are applied only during the test time on pretrained models and the results from different model outputs are combined to obtain the final prediction. While this means that no additional model training is needed, the tracking time increases linearly with the number of passes. In implicit inference, the dropouts are performed within the model, and feature maps resulting from different dropouts are combined together as a part of the model architecture. Thus, in implicit implementation of SD, the model is trained end-to-end to output a single prediction inferred over a set of dropouts in the latent space. }

The contributions are as follows.%ARN resumed here
\begin{itemize}
    \item We present three forms of structured dropout, to reduce the effect of occlusion in Siamese trackers which, different from other sources of variation, cannot be learned directly. 
    \item We present an end-to-end architecture, requiring a single pass for faster inference.
	\item Experiments show performance increase in a state-of-the-art Siamese tracker on various data sets (OTB2013, OTB2015, VOT2018, UAV123, LaSOT and \mbox{GOT-10k}).
	\
\end{itemize}

%% file: background.tex
\subsection{Siamese Tracking}

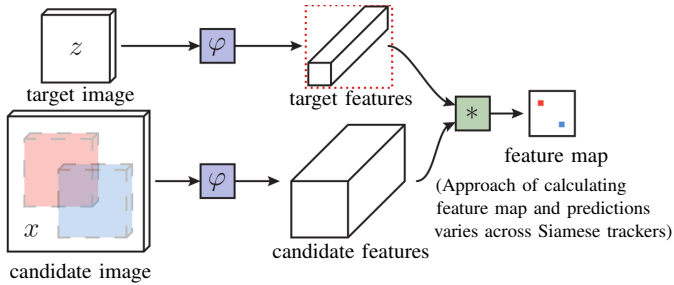
\begin{figure}
\centering
\begin{tikzpicture}
\node[inner sep=0pt] (img1) at (0,0)    {\includegraphics[scale=0.2]{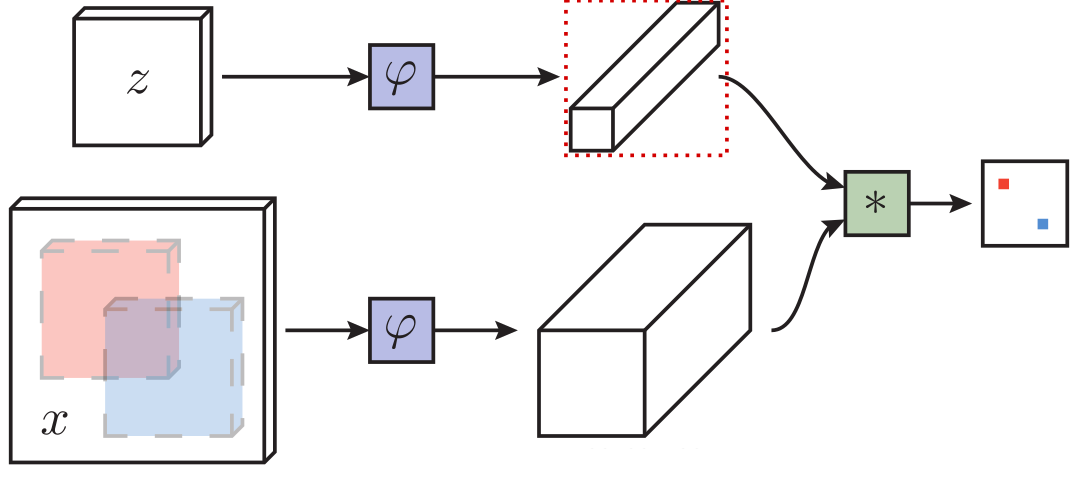}};
\node at (-2.76,-1.8) {\footnotesize candidate image};
\node at (-2.76, 0.55) {\footnotesize target image};
\node at (0.8, 0.5) {\footnotesize target features};
\node at (0.8,-1.5) {\footnotesize candidate features};
\node at (3.5, -0.25) {\footnotesize feature map};
\node at (3.17,-0.65) {\scriptsize (Approach of calculating};
\node at (3.33,-0.95) {\scriptsize feature map and predictions};
\node at (3.46,-1.25) {\scriptsize varies across Siamese trackers)};
%\node at (0,0) {\footnotesize feature map and final predictions varies across trackers}
\end{tikzpicture}
\caption{Schematic representation of SiamFC architecture (modified after \protect\cite{Bertinetto2016eccvw}).}
\label{fig_schem_siamfc}
\end{figure} 

We use SiamFC~\cite{Bertinetto2016eccvw} as a typical example of Siamese trackers. 
Siamese trackers use twin subnetworks that share weights, see Figure \ref{fig_schem_siamfc}.
The two subnetworks separately process the \emph{target} $z$ and \emph{candidate} $x$ images, see Figure \ref{fig_schem_siamfc}.
%The proposed structured dropouts are designed tailored to the structure of siamese trackers. Hence, before going into explaining the concept of structured dropouts, we provide an overview of the basic underlying concept of siamese trackers. Figure \ref{fig_schem_siamfc} shows the schematic structure of SiamFC architecture \cite{Bertinetto2016eccvw}. SiamFC is among the first Siamese models used in object tracking, and due the simplicity of architecture, we use it to provide a general overview of siamese trackers. This class of trackers use twin subnetworks that share weights, and the two subnetworks separately process the \emph{target} and \emph{candidate} images.} 
%Here, target refers to the object of interest as defined in the first frame of any sequence, and it needs to be tracked in the rest of the sequence. 

During tracking, the localization of $z$ in $x$ is found through matching in the latent space. The latent encodings $\Phi(z)$ and $\Phi(x)$ are obtained by the encoding subnetwork $\Phi(\cdot)$. Next, the two encodings are cross-correlated and combined into a single feature map, with the peak energy in the obtained feature map corresponding to the location of the target in $x$. Different Siamese trackers employ different mechanisms to localize the final bounding box output from these feature maps. SiamFC generates a feature map through matching with the target at several different scales, and the feature map pixel with highest value among those is translated back to the bounding box prediction. Other trackers (SiamRPN, DaSiamRPN, SiamRPN++) use a region proposal network to regress the desired bounding box. 

The latent encodings $\Phi(z)$ and $\Phi(x)$ comprise a set of N channels each. Each channel can be interpreted as a spatial map showing the distribution of a certain feature characteristic in the image. 
%Note that due to the nature of convolutional networks, the spatial information is preserved within each channel. 
Let $\tilde{z}$ denote the location of the variant of $z$ in $x$ that needs to be identified. For a case where $\tilde{z} \approx z$, we have $\Phi(\tilde{z}) \approx \Phi(z)$, and the cross-correlation map will show a high correlation score at the location corresponding to $\tilde{z}$. Similarly, for a case where $\tilde{z}$ is partially occluded, extent of correlation will reduce, since the occluded part will adversely affect the sum. To circumvent this issue, a possible solution would be to not consider the contribution from occluded parts of the image to the correlation score, and scale the score based on the rest. Contrary to how the matching is performed in the conventional Siamese networks, this approach would allow considering the occluded parts also as those of the target, without letting them impact the correlation score. 

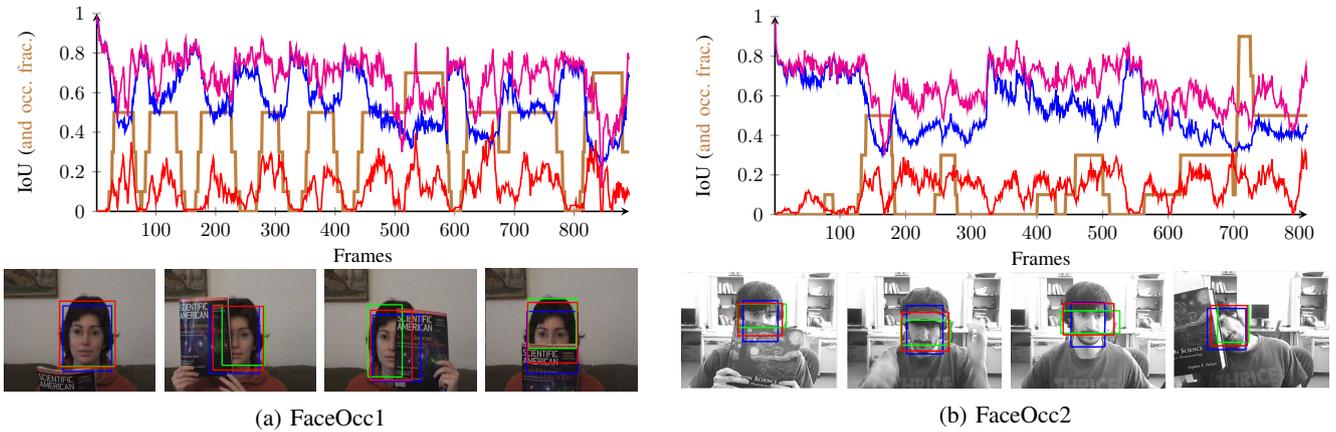
\begin{figure*}
\centering
	\begin{subfigure}{0.49\textwidth}
	\centering
	   	\begin{tikzpicture}[scale = 0.75]
		\begin{axis}[%
		width=4.3in,
		height=2in,
		thick, 
		axis x line=bottom,
		axis y line=left,
		legend style={at={(0.5,1.7)},anchor=north},
		%legend pos = outer north east,
		xlabel = {Frames},
		ylabel = {IoU ({\color{brown} and occ. frac.})},
		ymax = 1.0,
		ymin=0.0
		%ylabel style = {rotate = -90},
		%ymode = log
		]
		%\addlegendimage{empty legend}
		\addplot[mark=none, mark options={scale=1}, solid, brown,  ultra thick] table[x=fid, y = occfrac] {data/faceocc1.txt}; \label{line:iv}
		\addplot[mark=none, mark options={scale=1}, solid, blue,  thick] table[x=fid ,y = iou1]
		{data/faceocc1.txt}; \label{line:sv}
		\addplot[mark=none, mark options={scale=1}, magenta, thick] table[x=fid, y = iou2] {data/faceocc1.txt}; \label{line:occ}
		\addplot[mark=none, mark options={scale=1}, solid, red, thick] table[x=fid, y = diou2] {data/faceocc1.txt};\label{line:def}
%		\addplot[mark=square*, mark options={scale=1}, solid, blue, thick] table[x=rep,y = MB] {data_files/eco_auc.dat};\label{line:mb}
%		\addplot[mark=square*, mark options={scale=1}, solid, brown, thick] table[x=rep, y = FM] {data_files/eco_auc.dat};\label{line:fm}
%		\addplot[mark=*, mark options={scale=1}, solid, black, thick] table[x=rep, y = IPR] {data_files/eco_auc.dat};\label{line:ipr}
%		\addplot[mark=triangle*, mark options={scale=1}, dashdotted, blue, thick] table[x=rep, y = OPR] {data_files/eco_auc.dat};\label{line:opr}
%		\addplot[mark=*, mark options={scale=1}, solid, red, thick] table[x=rep, y = OV] {data_files/eco_auc.dat};\label{line:ov}
%		\addplot[mark=triangle*, mark options={scale=1}, dashdotted, red,  thick] table[x=rep, y = BC] {data_files/eco_auc.dat};\label{line:bc}
%		\addplot[mark=*, mark options={scale=1}, solid, cyan, thick] table[x=rep, y = LR] {data_files/eco_auc.dat};\label{line:lr}
% 		\addlegendentry{Illumination Variation}
% 		\addlegendentry{Scale Variation}
% 		\addlegendentry{Occlusion}
% 		\addlegendentry{Deformation}
		\end{axis}
		\end{tikzpicture}
% 		\vspace{-1em}
		\label{fig:difficulties-decay-sint}
\includegraphics[scale=0.16]{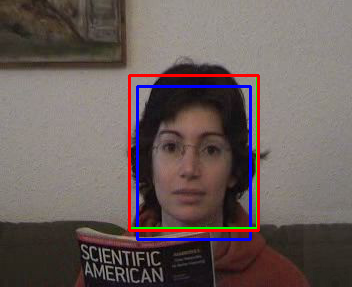}
\includegraphics[scale=0.16]{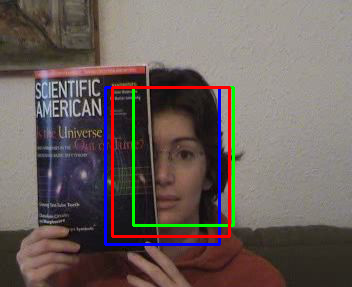}
\includegraphics[scale=0.16]{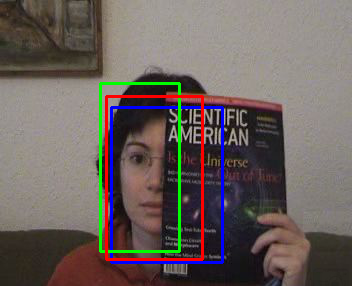}
\includegraphics[scale=0.16]{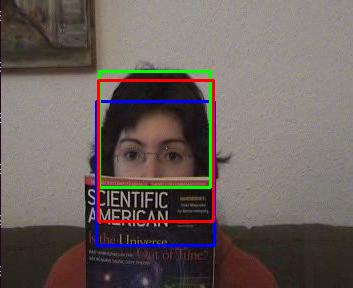}
\caption{FaceOcc1}
\end{subfigure}
	\begin{subfigure}{0.49\textwidth}
	\centering
	   	\begin{tikzpicture}[scale = 0.75]
		\begin{axis}[%
		width=4.3in,
		height=2in,
		thick, 
		axis x line=bottom,
		axis y line=left,
		legend style={at={(0.5,1.7)},anchor=north},
		%legend pos = outer north east,
		xlabel = {Frames},
		ylabel = {IoU ({\color{brown} and occ. frac.})},
		ymax = 1.0,
		ymin=0.0
		%ylabel style = {rotate = -90},
		%ymode = log
		]
		%\addlegendimage{empty legend}
		\addplot[mark=none, mark options={scale=1}, solid, brown,  ultra thick] table[x=fid, y = occfrac] {data/faceocc2.txt}; \label{line:iv}
		\addplot[mark=none, mark options={scale=1}, solid, blue,  thick] table[x=fid ,y = iou1]
		{data/faceocc2.txt}; \label{line:sv}
		\addplot[mark=none, mark options={scale=1}, magenta, thick] table[x=fid, y = iou2] {data/faceocc2.txt}; \label{line:occ}
		\addplot[mark=none, mark options={scale=1}, solid, red, thick] table[x=fid, y = diou2] {data/faceocc2.txt};\label{line:def}
%		\addplot[mark=square*, mark options={scale=1}, solid, blue, thick] table[x=rep,y = MB] {data_files/eco_auc.dat};\label{line:mb}
%		\addplot[mark=square*, mark options={scale=1}, solid, brown, thick] table[x=rep, y = FM] {data_files/eco_auc.dat};\label{line:fm}
%		\addplot[mark=*, mark options={scale=1}, solid, black, thick] table[x=rep, y = IPR] {data_files/eco_auc.dat};\label{line:ipr}
%		\addplot[mark=triangle*, mark options={scale=1}, dashdotted, blue, thick] table[x=rep, y = OPR] {data_files/eco_auc.dat};\label{line:opr}
%		\addplot[mark=*, mark options={scale=1}, solid, red, thick] table[x=rep, y = OV] {data_files/eco_auc.dat};\label{line:ov}
%		\addplot[mark=triangle*, mark options={scale=1}, dashdotted, red,  thick] table[x=rep, y = BC] {data_files/eco_auc.dat};\label{line:bc}
%		\addplot[mark=*, mark options={scale=1}, solid, cyan, thick] table[x=rep, y = LR] {data_files/eco_auc.dat};\label{line:lr}
% 		\addlegendentry{Illumination Variation}
% 		\addlegendentry{Scale Variation}
% 		\addlegendentry{Occlusion}
% 		\addlegendentry{Deformation}
		\end{axis}
		\end{tikzpicture}
% 		\vspace{-1em}
		\label{fig:difficulties-decay-sint}
\includegraphics[scale=0.18]{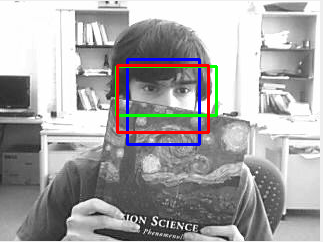}
\includegraphics[scale=0.18]{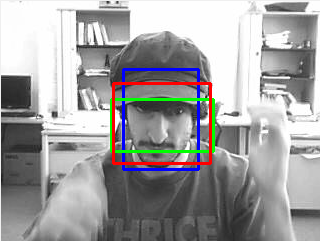}
\includegraphics[scale=0.18]{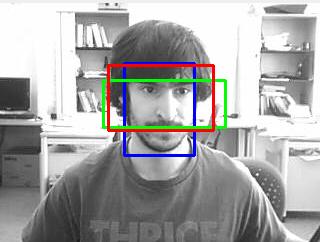}
\includegraphics[scale=0.18]{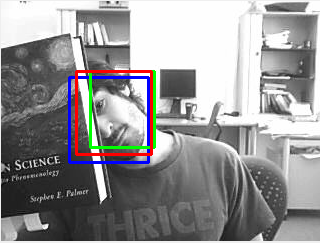}
\caption{FaceOcc2}
\end{subfigure}
\caption{IoU scores obtained using SiamRPN++ with (\ref{line:occ}) and without (\ref{line:sv}) (slice) structured dropouts, the change in IoU between the two methods (\ref{line:def}), and fraction of target occluded in every frame (\ref{line:iv}) of two video sequences from \mbox{OTB100 \cite{Wu2013cvpr}} dataset.}
\label{fig_schem_corr}
\end{figure*}

%\str{I would separate the sections. One for the siamese tracking, and maybe MC dropout. This should be a section called ``Background'' or ``Siamese tracking''. And then a new seciton called ``Structured dropouts''}.

\subsection{Monte Carlo dropouts}
Uncertainty in Bayesian neural networks can be decomposed into uncertainty associated with the model, miss-specifications in the model and inherent noise. Monte Carlo dropout is meant to tackle the first two of these three. The key idea is to use dropouts during inference in addition to the common use of dropout during learning. In \cite{Gal2016icml}, it has been demonstrated that using dropouts during inference is equivalent to performing Bayesian approximation by which process model uncertainty can be estimated. At test time, inference is performed multiple times with different dropouts, and mean output and prediction interval are then identified. For mathematical details related to this approach, see \cite{Gal2016icml}.

In \cite{Kraus2019arxiv}, uncertainty estimates obtained using MC-dropout have been shown to be weakly correlated with occlusion. Further, in \cite{Miller2018icra}, it has been demonstrated that an ensemble estimate over multiple predictions is closer to the ground-truth. However, due to the stochastic nature of the approach, large number of samples are needed to be able to make a good inference reducing the inference speed drastically, unfeasible for tracking.

%% file: propose_dropouts.tex
\subsection{Description}

\emph{Structured dropout} refers to dropout applied in the latent space of the target subnetwork of siamese trackers to mimic target occlusion. Here and henceforth, this term will be interchangeably used to refer to the dropout mechanism as well as the siamese tracker that is modified with it.
Example demonstrations of improvements obtained through aggregation of multiple structured dropouts are shown in Figure \ref{fig_schem_corr}. It can be seen that for several instances for the two sequences, using structured dropouts provides a better localization of the bounding boxes, except for rare instances where it degrades slightly. Most importantly, we observe that the gain is mostly from frames where occlusion is prominent. While using the structured dropouts does not completely handle the challenge of occlusion, the obtained improvements demonstrate a certain degree of correlation between the two.

We propose different variants of structured dropouts based on the types of occlusion. We identify two types of occlusion in object tracking: \emph{feature occlusion} and \emph{patch occlusion}. Feature occlusion refers to occluded parts in the image arising from changes in the target, which lead to the disappearance of some of the characteristic features of the target while most others still exist. Feature occlusion includes for example, images of a person with sunglasses or a raincoat obscuring part of the original target features. A modern tracker might be able to handle some instances of such occlusions on the basis of discrimination with missing or false information but only to a certain degree. 
%Feature occlusion is therefore reinforced through structured occlusion.

% \begin{figure}
% 	\centering
% 	\begin{subfigure}{0.48\textwidth}
% 		\centering
% 		\includegraphics[height=1.8cm, width=2.5cm]{images/otb_samples/Girl_1.png}
% 		\includegraphics[height=1.8cm, width=2.5cm]{images/otb_samples/Girl_2.png}
% 		\caption{{\color{red} Add conf. score}}
% 		\label{fig:occ_case1}
% 	\end{subfigure}\hspace{-1em}
% 	\begin{subfigure}{0.48\textwidth}
% 		\centering
% 		\includegraphics[height=1.8cm, width=2.5cm]{images/otb_samples/FaceOcc1_1.png}
% 		\includegraphics[height=1.8cm, width=2.5cm]{images/otb_samples/FaceOcc1_2.png}
% 		\caption{{\color{red} Add conf. score}}
% 		\label{fig:occ_case2}
% 	\end{subfigure}
% 	\caption{Example frames from two videos of OTB100 dataset \protect\cite{} demonstrating the effect of occlusion on predictions of SiamRPN++ tracker \protect\cite{}.}
% 	\label{fig:occ_ex1}
% \end{figure}
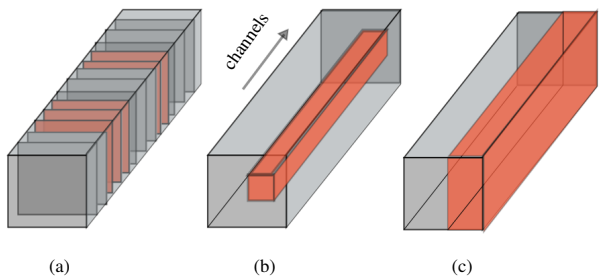
\begin{figure}
	\centering
	
	\begin{tikzpicture}[scale = 1]
	%\addlegendimage{empty legend}
	\draw (0, 0) node[inner sep=0]{\includegraphics[scale=0.25]{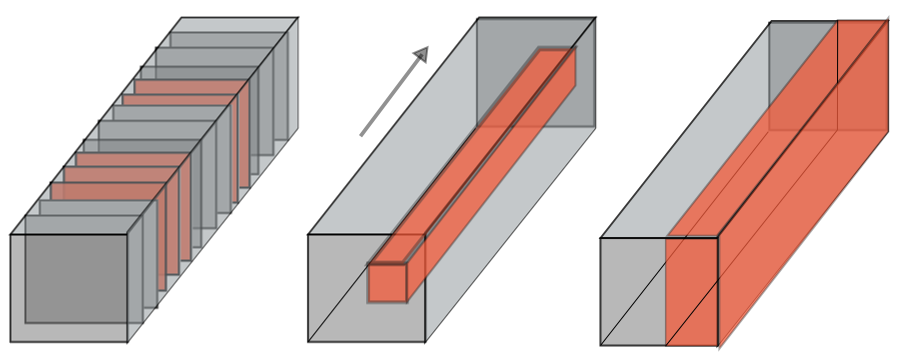}};
	\draw (-0.75, 1) node[rotate=55, scale=0.7] {channels};
	\draw (-3.2, -2) node[scale=0.7] {(a)};
	\draw (-0.5, -2) node[scale=0.7] {(b)};
	\draw (2.1, -2) node[scale=0.7] {(c)};
	\end{tikzpicture}
	\vspace{-0.5em}
	\caption{Schematic representation of three structured dropout strategies for handling occlusion: (a) channel dropout, (b) segment dropout, and (c) slice dropout.}
	\label{fig:dropout_ex1}
\end{figure} 
Patch occlusion refers to compact parts of the target being blocked by another object. Generally, patch occlusions occur when another object enter the target's view from one of the sides and partly blocks the view of the target, see Figure \ref{fig_schem1}. 

To tackle the two types of occlusion problems described above, we propose three structured dropout methods: \emph{channel}, \emph{segment} and \emph{slice dropout} (see Figure \ref{fig:dropout_ex1}) applied on the latent feature map of the target $\Phi(z)$ obtained from the target subnetwork, as shown in Figure \ref{fig_schem_siamfc}.
%{\color{green} ARN I donot understand the next two sentences.} The latent feature  maps for the two subnetworks denote the spatial distribution of the set of features in the two input images. These features, not necessarily independent of one another, define the characteristics of the inputs. 

\textbf{Channel dropout. } These are designed to specifically handle feature occlusions. 
When applying channel dropouts, a randomly selected set of channels from $\Phi(z)$ are set to 0. Setting parts of $\Phi(z)$ to 0 implies that while matching, the corresponding features of the template image will not contribute. This process of dropping out a random set of channels is repeated $n$ times and the results are combined to obtain the final prediction.

\textbf{Segment dropout. } To handle patch occlusions, we propose to match the target and the candidate boxes by dropping parts of the latent feature maps along the spatial dimension of the template subnetwork. As shown in Figure \ref{fig:dropout_ex1}, segment dropout involves random dropping off part of the feature map across all the channels in the target subnetwork, thereby ignoring the information contained in it during the matching process. 

\textbf{Slice dropout. } Since occlusion can stochastically occur in any part of the target, segment dropouts are also stochastic. However, this also means that to match optimally to the occluded target, a large number of segment dropouts would normally be required, thus increasing the computational footprint of the method. We note that a specific and frequently occurring case of patch occlusion is that the target gets occluded on one of the sides, be it the left, right, bottom or top. For example, pedestrians mostly occlude each other from either left or right. Therefore, instead of sampling all random locations, we iterate over a predefined set of occlusion patches sampled from the different sides of the target, thus covering most cases of occlusion and achieving a good trade-off with complexity. We refer this approach as slice dropout. Example results where slice dropouts worked well are shown in \mbox{Figure \ref{fig_schem_corr}.}

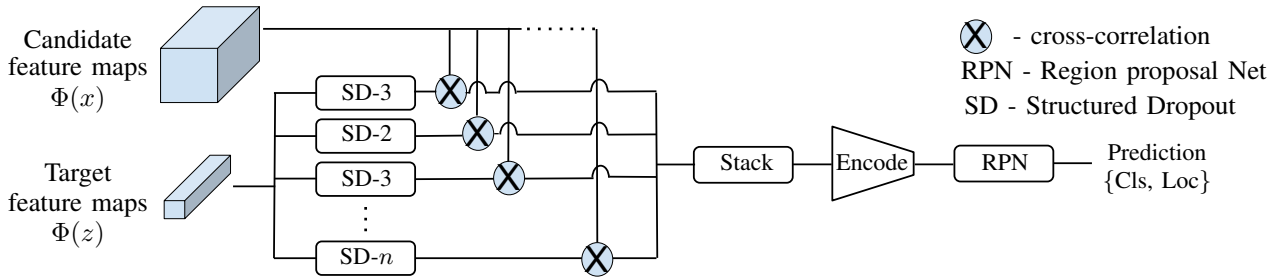
\begin{figure*}
\centering
% \begin{subfigure}{1.0\textwidth}
% \centering
% \begin{tikzpicture}
% \node (img) {\includegraphics[scale=0.45, trim= 0cm 8cm 3cm 0, clip]{images/concepts/fig_schem_exp.pdf}};
% \node[] at (6.3,-0.4) {\small Prediction};
% \node[] at (6.3,-0.8) {\small \{Cls, Loc\}};
% \node[] at (4.2,-0.62) {\small Ensemble};
% \node[] at (1.85,-0.5) {\small Pred-3};
% \node[] at (1.85, 0.05) {\small Pred-2};
% \node[] at (1.85, 0.6) {\small Pred-1};
% \node[] at (1.85, -1.55) {\small Pred-$n$};
% \node[] at (0.3,-0.5) {\small RPN};
% \node[] at (0.3, 0.05) {\small RPN};
% \node[] at (0.3, 0.6) {\small RPN};
% \node[] at (0.3, -1.58) {\small RPN};
% \node[] at (-4.2,-0.54) {\small SD-3};
% \node[] at (-4.2, 0.05) {\small SD-2};
% \node[] at (-4.2, 0.6) {\small SD-3};
% \node[] at (-4.2, -1.58) {\small SD-$n$};
% \node[] at (-8.0, 1.3) {Candidate};
% \node[] at (-8.0, 0.9) {feature maps};
% \node[] at (-8.0, 0.5) {$\Phi(x)$};
% \node[] at (-8.0, -0.5) {Target};
% \node[] at (-8.0, -0.9) {feature maps};
% \node[] at (-8.0, -1.3) {$\Phi(z)$};
% \node[] at (5.6, 1.35) { - cross-correlation};
% \node[] at (5.63, 0.85) {RPN - Region proposal Net};
% \node[] at (5.45, 0.38) {SD - Structured Dropout};
% \end{tikzpicture}
% \vspace{-2em}
% \caption{Explicit}
% \end{subfigure}
\begin{subfigure}{1.0\textwidth}
\vspace{-2em}
\centering
\begin{tikzpicture}
\node (img) {\includegraphics[scale=0.45, trim= 0cm 8cm 3cm 0, clip]{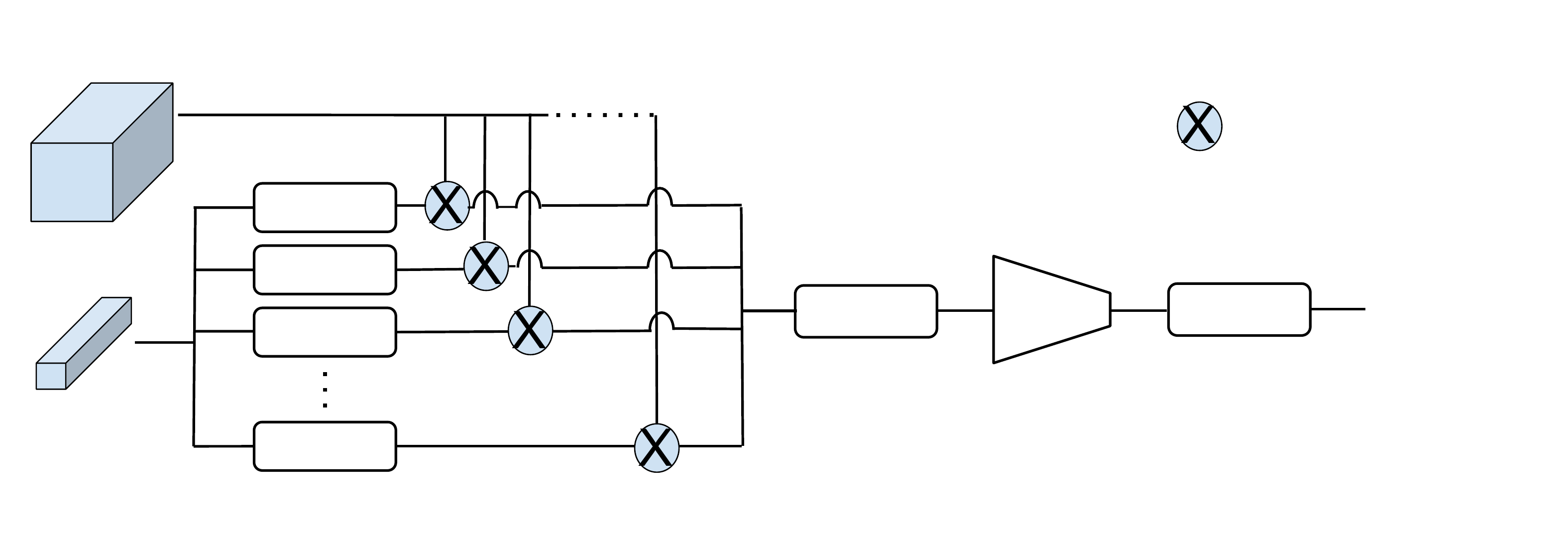}};
\node[] at (6.2,-0.2) {\small Prediction};
\node[] at (6.2,-0.6) {\small \{Cls, Loc\}};
\node[] at (4.2,-0.33) {\small RPN};
\node[] at (2.45,-0.33) {\small Encode};
\node[] at (0.8, -0.33) {\small Stack};
\node[] at (-4.2,-0.54) {\small SD-3};
\node[] at (-4.2, 0.05) {\small SD-2};
\node[] at (-4.2, 0.6) {\small SD-3};
\node[] at (-4.2, -1.58) {\small SD-$n$};
\node[] at (-8.0, 1.3) {Candidate};
\node[] at (-8.0, 0.9) {feature maps};
\node[] at (-8.0, 0.5) {$\Phi(x)$};
\node[] at (-8.0, -0.5) {Target};
\node[] at (-8.0, -0.9) {feature maps};
\node[] at (-8.0, -1.3) {$\Phi(z)$};
\node[] at (5.6, 1.35) { - cross-correlation};
\node[] at (5.63, 0.85) {RPN - Region proposal Net};
\node[] at (5.45, 0.38) {SD - Structured Dropout};
\end{tikzpicture}
\vspace{-2em}
%\caption{Implicit}
\end{subfigure}
\caption{Schematic representations of inference procedure for our structured dropouts. The model architecture is trained end-to-end, thereby requiring only a single pass during inference.}
\label{ref_fig_schem_infer}
\end{figure*}

\subsection{Final Prediction from Dropout Sampling}
For the structured dropout procedures presented above, the outputs from multiple passes are combined to obtain the final prediction. A simple and straightforward approach is to combine multiple predictions using a non-max suppression approach. This can be interpreted similar to the way results for MC-Dropout are combined over a number of passes, as demonstrated in \cite{Kraus2019arxiv}. An advantage of such a regime is that pretrained tracker models can directly be used with only the addition of the structured dropouts. However, such an approach provided only limited improvement in tracker performance for channel dropouts. Moreover, this approach is sensitive to the choice of hyperparameters, and adversely affected the tracker performance for  segment and slice dropouts. 

Rather than combining multiple dropout predictions explicitly, we directly combine the activation maps using a small encoder network, and the modified model with structured dropouts is trained in a end-to-end manner to produce the final prediction. Schematic representation of this architecture is shown in Figure \ref{ref_fig_schem_infer}. The $n$ feature map samples obtained from $n$ dropouts are stacked and passed through a set of convolutional layers, and the obtained output is then passed into the regression module (\emph{e.g.}, regional proposal network module in SiamRPN and SiamRPN++) similar to the conventional approach of Siamese tracking models. 

%Since implicit implementation involves additional convolutions applied on $\Psi_{z,x}$, the choice of these added convolutions depends on the type of dropouts. For example, slice dropouts are non-stochastic, and positional dependence of the each of the dropouts can be learned in the trained model. This implies that if trained on a diverse dataset with different scenarios of occlusions, the additional layers should learn the probabilities of occurrence of occlusion at the different edges of the target. With the series of convolutions, we expect the model to learn these positional values, be it that there are only limited cases of occlusion in our data. On the other hand, positional dependence cannot be learned for channel and segment dropouts due to their stochastic nature. However, the added convolutions are expected to learn operations applicable over a set. 

%% file: related_work.tex
Since this paper aims at proposing a solution for tackling occlusion in siamese trackers, we first discuss a few recent trackers, with a special focus on siamese trackers. The field of visual tracking has evolved rapidly in the recent years, due to the development of several new benchmark datasets \cite{Wu2013cvpr, Wu2015cvpr, Muller2016eccv, Huang2020tpami, Fan2019cvpr, Kristan2018eccvw}, and improved tracking models \cite{Danelljan2015iccv, Danelljan2016eccv, Danelljan2017cvpr, Bhat2019iccv, Nam2016cvpr}. Recently, with the adoption of deep learning in visual object tracking, tracking models using correlation filter combined with deep feature representations have obtained state-of-art accuracies on most tracking benchmarks \cite{Danelljan2017cvpr, Wu2013cvpr, Wu2015cvpr, Huang2020tpami, Li2018cvpr, Li2019cvpr, Bhat2019iccv}. Among these, most trackers rely on the siamese matching backbone, and these trackers have received significant attention due to their well-balanced accuracy and efficiency. We design solution for tackling occlusion especially tailored for siamese trackers, and further below, we discuss this class of trackers.

\textbf{Siamese Tracking. } These trackers pose object tracking as a problem of cross-correlation amd the models can be trained end-to-end in a deep learning setting. The first siamese trackers~\cite{Tao2016cvpr, Bertinetto2016eccvw} relied on a tracking by similarity comparison strategy. These trackers simply search for the candidate most similar to the original image patch of the target given in the starting frame, using a run-time fixed but learned \textit{a priori} deep siamese similarity function. Due to their no-updating nature, siamese trackers are robust against several sources of error that can cause tracker drift. 

Improved versions of siamese trackers include regional proposal frameworks (\emph{e.g.} SiamRPN \cite{Li2018cvpr}, DaSiamRPN \cite{Zhu2018eccv}) and deeper backbones such as ResNet50 (SiamDW \cite{Zhang2019cvpr}, SiamRPN++ \cite{Li2019cvpr}, DCFST \cite{Zheng2019aaai}, among others) which provide more accurate predictions and improve the tracking performance. Due to the weight sharing in the twin subnetworks, we argue that the effect of occlusion can be mimiced in the latent space of siamese trackers through customized dropout strategies.

%A seminal work on long-term tracking is the TLD~\cite{kalal2012tracking} tracker. It is a multi-component method that combines an optical flow tracker with a detector, and the detector part is slowly updated. However, TLD is very sensitive to tracking circumstances. SPL~\cite{supancic2013self} follows the TLD paradigm for long-term tracking, and employs a SVM-based detector which is updated using the frames that produce the lowest SVM-objective for the training set. The repeated evaluation of the SVM objective, however, has a noticeable computational footprint.

%LTCT~\cite{MaYZY15} is a correlation filter tracker paying particular attention to long-term tracking, by integrating an online detector to re-detect the target in case of tracking failures. Furthermore, in the recent long-term challenge by VOT several new trackers were proposed~\cite{Kristan2018a}, grouped into four families: short-trackers that do not implement re-detection nor model occlusion, short-term trackers with conservative updating, pseudo  long-term trackers that do not return a box when the target is not visible, and re-detecting long-term trackers.

\textbf{Tackling Occlusion. } Occlusion is considered among the most difficult challenges in the field of computer vision, and especially in object tracking. Occlusion detection has been applied to improve the performance of various tasks in computer vision, \emph{e.g.}, action recognition \cite{Weinland2010eccv}, 3D reconstruction \cite{Schonberger2016cvpr}, among others. To our knowledge, only limited works exist that focus on handling this problem in the context of object tracking, and these do not exploit the capability of deep learning models. For example, in \cite{Pan2007cvpr}, authors progressively analyze occlusion situation during the course of tracking through exploiting the spatiotemporal context information. Most of the methods defined to tackle occlusion either in tracking or other problems in computer vision iteratively refine their estimate on occlusion through progressively improving their motion accuracy. 

In \cite{Wang2018cvpr}, deep learning is employed and occlusion is inferred through an end-to-end trainable motion estimation network. However, such a solution for occlusion detection would be prone to model decay \cite{Gavves2019arxiv}, caused due to the tracker drift. To tackle this issue, a solution for occlusion detection is needed which can treat every frame of a video in an independent manner, thus not relying on motion information. In \cite{li2018bmvc}, extent of occlusion is estimated using siamese networks, however, the proposed approach is meant for occlusion encountered in binocular vision, where only minor occlusions occur between a pair of images with the rest almost unchanged. Similar to the work of \cite{li2018bmvc}, we eliminate the need for motion information and use siamese network to independently identify occlusion scenario in every frame of a video. However, we achieve this mimicing of occlusion through the use of structured dropouts in the latent space of the model.

%% file: tracker_choice.tex
\textbf{Base Siamese Trackers. }Among the various object tracking algorithms based on Siamese matching, we focus on two implementations: \mbox{SiamFC \cite{Bertinetto2016eccvw}} and SiamRPN++ \cite{Li2019cvpr}. Due to its fully-convolutional nature, SiamFC is memory-efficient as well as provides high inference speed. Due to these characteristics, it forms the backbone for almost all recent siamese trackers. The simplicity of SiamFC implementation makes it a suited candidate to studied in the context of structured dropouts proposed in this study. SiamRPN++ is chosen with the intention of demonstrating that the recent, best performing siamese trackers can also be improved using structured dropouts. This tracker is among the state-of-the-art tracking algorithms. 
%\ssout{, with only DiMP \cite{} performs better, as we know of. DiMP uses an online learning component as well as an optimization module in the target subnetwork, due to which, the implementation of structured dropouts is not very straightforward. On the contrary, SiamRPN++ uses a plain implementation of siamese matching, and is thus preferred as a second tracking algorithm for this study.}

\textbf{Explicit Dropout Sampling. }As stated earlier, the results from multiple passes of structured dropouts need to be combined to obtain the final prediction. As a baseline method, we follow a sampling approach similar to that described in \cite{Miller2018icra}. The corresponding models are referred later in this paper with the prefix `exp-'. Let the prediction by the tracking model for the $i^\text{th}$ model pass be denoted by $\mathbf{D}_{t,i} = \{\mathbf{y}_{t,i}, s_{t,i}\}$, where $\mathbf{y}_{t,i}$ and $s_{t,i}$ denote the predicted bounding box and the respective confidence score for the $i^{\text{th}}$ forward pass at time step $t$. From $n$ forward passes, the set of predictions can then be denoted as $\mathcal{D}_t = \{\mathbf{D}_{t,1}, \mathbf{D}_{t,2},\hdots, \mathbf{D}_{t,n}\}$.

For channel dropouts, the observation set $\mathcal{D}_t$ is partitioned into a multiple subsets based on mutual \emph{Intersection-Over-Union (IoU)} scores, such that for any two observations $\{\mathbf{D}_{t,j}, \mathbf{D}_{t,k}\}$ drawn from a certain subset, \mbox{$IoU(\mathbf{y}_{t,i}, \mathbf{y}_{t,k}) > \alpha_c$}. From the susbset containing maximum observations, the one with highest confidence score is chosen as the final prediction. For segment or slice dropout, let the area fraction of the dropout segment for the $i^{\text{th}}$ pass at step $t$ be denoted by $\mathcal{A}_{t,i}$. The confidence score for this box is then scaled up as $s^*_{t,i} = \frac{\mathcal{B}}{1-\mathcal{A}_{t,i}}\cdot s_{t,i}$. This is done to balance the drop in confidence score resulting from dropping out the patch. The constant $\mathcal{B}$ is added to control the impact of upscaling on the confidence scores. After upscaling, the bounding box with the highest value of $s^*_{t}$ is chosen as the final prediction for the current tracking step. For experiments reported in this paper, $\alpha_c$ and $\mathcal{B}$ are set to 0.2 and 0.9, respectively.

%% file: data.tex
\textbf{Training.} The SiamFC model and its variant with structured dropouts are trained on the training set of GOT-10k dataset \cite{Huang2020tpami}. For training SiamRPN++ \cite{Li2019cvpr}, the training set of COCO \cite{Lin2014eccv}, ImageNet DET \cite{Russakovsky2015ijcv}, ImageNet VID and YouTube-BoundingBoxes Dataset \cite{Real2017cvpr} were used, with all training parameters set similar to that specified in \cite{Li2019cvpr}. 

\textbf{Evaluation.} The focus of this paper is to improve for occlusion-effects in short-term tracking, and the performance of structured dropouts is demonstrated on OTB2015 \cite{Wu2015cvpr}, VOT2018 \cite{Kristan2018eccvw}, UAV123 \cite{Muller2016eccv}, GOT-10k test set \cite{Huang2020tpami} and LaSOT \cite{Fan2019cvpr}. 

%% file: imp_details.tex
%\str{Can you organize this per architecture. Like,
%SiamFC-SD ... And SiamRPN-SD. SD stands for structured dropouts.}

\textbf{Model Architecture. }The architectural modifications associated with structured dropouts depend on the choice of the base siamese tracker. For SiamFC and SiamRPN++, the new architectures (with structural dropouts) will be referred as SiamFC-SD and \mbox{SiamRPN-SD}. For both the implementations, number of passes $n$ is chosen as 21 for channel and segment dropouts. For slice dropout, $n$ is set to 13 and 9 for SiamFC-SD and SiamRPN-SD, respectively. For all the cases, one of the total passes corresponds to no dropout applied. Random dropouts of 0.2 are chosen for channel and segment dropouts. Further details specific to the two implementations follow below. 

\emph{SiamFC-SD. }An encoder architecture is added  immediately after the cross-correlation module to combine the sets of activation maps from multiple passes of the dropouts. It comprises two convolutional layers that combine the activation maps obtained from multiple passes of dropouts and produce a new map that can be used by the later part of the model architecture for target localization. SiamFC uses a cross-correlation layer to obtain a single channel response map for localizing the target. Thus, the added convolutions need to combine the $n$ activation maps in this case. For SiamFC, applying structured dropouts results in 21 and 13 channels in the activation map for channel and segment dropouts and slice dropout, respectively. With kernel and group sizes set to 1 for the convolutional layers, these are projected to 4 channels and then to an activation map with a single channel. Both the convolutional layers are succeeded by a batch normalization and a that is used to regress the location of the target, as in the original SiamFC tracker \cite{Bertinetto2016eccvw}.

\emph{SiamRPN-SD. } For SiamRPN++ with channel or segment dropout, the resultant activation maps obtained from performing depthwise cross-correlations are passed through two convolutional layers. For all the cases, the activations without any dropout are also included. The dropouts are applied across all the 3 feature maps in a similar manner. The depthwise cross-correlation and layerwise aggregation module produces 3 sets of activation maps, each containing $n \times N_c$ channels, where $N_c(=$256) denotes the number of channels contained in the cross-correlated map before applying dropouts. The 21 variants of every channel are mapped on to 4 and then to 1. With two convolutional layers, the aggregated activation map comprising 5376 channels gets mapped to 1024 channels and then to 256 channels. The convolutions are performed in a similar manner reducing the number of channels from 2304 to 1024 and then to 256. The output after two convolutions is then fed back to the next layer from the original region proposal module of the tracker.

\textbf{Optimization.} All network implementations presented in this paper have been trained with stochastic gradient descent method. The training procedures for the baseline SiamFC model as well as our method are kept to be the same as those mentioned in \cite{Bertinetto2016eccvw}, and the model is trained on a machine equiped with a single NVIDIA GeoForce GTX Titan X. For SiamRPN++ baseline, procedure similar to \cite{Li2019cvpr} is followed and four GPUs with training minibatches of 16 are used.

%% file: results.tex
%-----SiamFC-----------------

This section presents an analysis on the results performed on the two baseline siamese trackers combined with the proposed structured dropout strategies (SiamFC-SD and SiamRPN-SD). We primarily compare the results of structured dropouts with that of the base siamese trackers without dropouts (SiamFC and SiamRPN), the ones with MC-dropout (denoted with suffix `-MC' and dropout strategies implemented with explicit sampling (denoted with prefix `exp-')).

\textbf{SiamFC with Structured Dropouts.} We study the performance of SiamFC-SD on OTB2015, UAV123, GOT-10k and VOT2018. Details on the results for the three structured dropout strategies follow below.
%The results related to the structured dropout implementations over SiamFC tracker are shown in Table \ref{table_results_siamfc}. Here, baseline refers to the original tracking model without any structured dropout implementation. Further, we also present results related to a basic MC-dropout implementation where 20\% of the channels are randomly dropped in total across the outputs from the last three convolutional layers of the model. 
\begin{table*}
\footnotesize
	\begin{center}
		\begin{tabular}{ |l || c|c|| c|c|| c |c |c||c|c|} 
			\hline
			 & \multicolumn{2}{c||}{OTB2015} & \multicolumn{2}{c||}{UAV123} & \multicolumn{3}{c|}{GOT-10k} &
			 \multicolumn{2}{c|}{VOT2018}\\
			Approach & Pr & SR$_{0.5}$ & Pr & SR$_{0.5}$ & AO & SR$_{0.5}$ & SR$_{0.75}$ & EAO & Acc \\ \hline 
			\hline
			SiamFC &  \textbf{0.809} & 0.597 & 0.711 & 0.513 & 0.355 & 0.395 & 0.118 & \textbf{0.311} & \textbf{0.508} \\
			MC-Dropout & 0.807 & 0.602 & 0.712 & 0.515 & 0.350 & 0.396 & 0.116 & 0.307 & 0.506 \\
			exp-SiamFC-SD & 0.807 & 0.608 & 0.732 & 0.526 & \textbf{0.366} & \textbf{0.409} & \textbf{0.131} & \textbf{0.311} & 0.506 \\
			SiamFC-SD (Ours)  & 0.808 & \textbf{0.610} & \textbf{0.736} & \textbf{0.535} & 0.361 & 0.402 & 0.129 & 0.309 & \textbf{0.508} \\
			\hline
		\end{tabular}
	\end{center}
	\caption{Performance values for SiamFC \protect\cite{Bertinetto2016eccvw} method with and without channel dropouts. }
	\label{table_results_siamfc1}
\end{table*} 

\emph{Channel SiamFC-SD.} From Table \ref{table_results_siamfc1}, we see that channel dropouts bring improvement consistently across all the 3 datasets, except VOT2018. On UAV123 dataset, absolute improvements of 2.6\% and 2.2\% are obtained on precision and success scores, respectively. For GOT-10k dataset, we observe that while the structured dropouts help, the explicit sampling baseline performs slightly better than our proposed model architecture. This implies that the process of combining the activations obtained from multiple dropouts is not sufficiently captured by our proposed encoding architecture, and further improving its complexity could possibly help.

\begin{table*}
\footnotesize
	\begin{center}
		\begin{tabular}{ |l || c|c|| c|c|| c |c |c||c|c|} 
			\hline
			 & \multicolumn{2}{c||}{OTB2015} & \multicolumn{2}{c||}{UAV123} & \multicolumn{3}{c|}{GOT-10k} &
			 \multicolumn{2}{c|}{VOT2018}\\
			Approach & Pr & SR$_{0.5}$ & Pr & SR$_{0.5}$ & AO & SR$_{0.5}$ & SR$_{0.75}$ & EAO & Acc \\ \hline 
			\hline
			SiamFC &  \textbf{0.809} & 0.597 & 0.711 & 0.513 & 0.355 & 0.395 & 0.118 & 0.311 & 0.508 \\
			MC-Dropout & 0.807 & 0.602 & 0.712 & 0.515 & 0.350 & 0.396 & 0.116 & 0.307 & 0.506 \\
			exp-SiamFC-SD & 0.801 & 0.591 & 0.692 & 0.499 & 0.348 & 0.387 & 0.117 & 0.277 & 0.472 \\
			SiamFC-SD (Ours) & 0.805 & \textbf{0.613} & \textbf{0.733} & \textbf{0.529} & \textbf{0.359} & \textbf{0.412} & \textbf{0.126} & \textbf{0.314} & \textbf{0.512} \\
			\hline
			
		\end{tabular}
	\end{center}
	\caption{Performance values for SiamFC \protect\cite{Bertinetto2016eccvw} method with and without segment dropouts.}
	\label{table_results_siamfc2}
\end{table*} 

\emph{Segment SiamFC-SD. } Table \ref{table_results_siamfc2} presents results related to SiamFC-SD with segment dropouts. Our approach improves performance values for all the 4 datasets. Improvements on VOT2018 are very small. Similar to channel droputs, approximate improvements of 2\% are obtained in prediction as well as success scores on UAV123. We observe noticeable drop in performance values for exp-SiamFC-SD with segment dropouts. The reason for this corresponds to the fixed choice of $\mathcal{B}$ across all sequences. We have observed that performance of segment dropouts on video sequences is sensitive to the choice of $\mathcal{B}$, and the value of 0.9 was obtained through iterative optimization on a subset of sequences.

\begin{table*}
\footnotesize
	\begin{center}
		\begin{tabular}{ |l || c|c|| c|c|| c |c |c||c|c|} 
			\hline
			 & \multicolumn{2}{c||}{OTB2015} & \multicolumn{2}{c||}{UAV123} & \multicolumn{3}{c|}{GOT-10k} &
			 \multicolumn{2}{c|}{VOT2018}\\
			Approach & Pr & SR$_{0.5}$ & Pr & SR$_{0.5}$ & AO & SR$_{0.5}$ & SR$_{0.75}$ & EAO & Acc \\ \hline 
			\hline
			SiamFC &  \textbf{0.809} & 0.597 & 0.711 & 0.513 & 0.355 & 0.395 & 0.118 & 0.311 & 0.508 \\
			MC-Dropout & 0.807 & 0.602 & 0.712 & 0.515 & 0.350 & 0.396 & 0.116 & 0.307 & 0.506 \\
            exp-SiamFC-SD & 0.794 & 0.588 & 0.706 & 0.505 & 0.345 & 0.387 & 0.104 & 0.301 & 0.495 \\
			SiamFC-SD (Ours) & 0.806 & \textbf{0.616} & \textbf{0.743} & \textbf{0.534} & \textbf{0.368} & \textbf{0.411} & \textbf{0.132} & \textbf{0.310} & \textbf{0.513} \\
			\hline
		\end{tabular}
	\end{center}
	\caption{Performance values for SiamFC \protect\cite{Bertinetto2016eccvw} method with and without slice dropouts.}
	\label{table_results_siamfc3}
\end{table*}

\emph{Slice SiamFC-SD. } The performance scores for the 4 datasets obtained using slice dropouts with SiamFC are stated in Table \ref{table_results_siamfc3}. The observations are mostly similar to those of segment dropouts. In particular, we observe that the performance improvements obtained using slice dropouts are slightly larger for most cases. This implies that the predefined dropout regions specified in slice dropouts help better in identifying the target. The explicit implementations for slice dropouts also report reduced performance, and the reason is similar to that of segment dropouts, as stated above.

 A general observation across all datasets for the three dropout strategies is that the conventional MC-dropout approach with an equivalent amount of computational budget as exp-SiamFC-SD brings in no improvement in tracker performance. From the observations stated above, we deduce that structured dropouts could potentially help in improving tracker's performance. Based on this motivation, we further explore their applicability with SiamRPN++, a state-of-the-art siamese tracker. Also, since explicit sampling combined with segment and slice dropouts degrade tracker performance, we do not consider these implementations in the further experiments of this paper.

%-----SiamRPN++-----------------

\begin{table}
\footnotesize
	\begin{center}
		\begin{tabular}{ |l || c|c||c|c|} 
			\hline
			 & \multicolumn{2}{c||}{OTB2015} &
			 \multicolumn{2}{c|}{VOT2018} \\
			Approach & Pr & Acc & EAO & Acc \\ \hline 
			\hline
			SiamRPN++ &  0.890 & 0.683 & 0.414 & 0.600 \\
%			SiamRPN++ \cite{Li2019cvpr} & 0.914 & 0.696 & 0.807 & 0.613 & - & - & - & - & - & 0.496 & - & 0.569 \\
			DiMP-50 \cite{Bhat2019iccv} & - & 0.684 &  0.440 & 0.597 \\
			UPDT \cite{Bhat2018eccv} & - & 0.702  &  0.378 & 0.536  \\
			ATOM \cite{Danelljan2019cvpr} & - & 0.669 &  0.401 & 0.590 \\
			SiamRPN-MC & 0.876 & 0.681 &  0.417 & 0.599 \\
			exp-SiamRPN-SD-channel & 0.908 & 0.695 &  0.416 & 0.591 \\
%			Exp-Segment & 0.858 & 0.672 \\
%			Exp-Slice &  \\
			%Exp-Rotate & -& -& -& -& -& -&- \\
			SiamRPN-SD-channel  & 0.912 & 0.702 &  0.421 & 0.601 \\
			SiamRPN-SD-segment & 0.896 & 0.698 & 0.410 & 0.588 \\
			SiamRPN-SD-slice & 0.914 & 0.701 & 0.418 & 0.598\\
			%Imp-Rotate & 0.807 & 0.613 & 0.528 & 0.721 & 0.380 & 0.425 & 0.140 \\
			%Imp-Slice-Rot & \textbf{0.811} & 0.614 & 0.533 & \textbf{0.745} & \textbf{0.388} & \textbf{0.429} & \textbf{0.146} \\
			\hline
			
		\end{tabular}
	\end{center}
	\caption{Performance values for SiamRPN++ \protect\cite{Bertinetto2016eccvw} with and without structured dropouts (SiamRPN-SD). The additions `MC' and `exp-' correspond to MC dropout and explicit SD sampling, respectively.}
	\label{table_results_siamrpnpp}
\end{table}

\begin{figure}
\centering
\begin{subfigure}{0.49\textwidth}
\centering
\includegraphics[height=1.9cm, width=2.8cm]{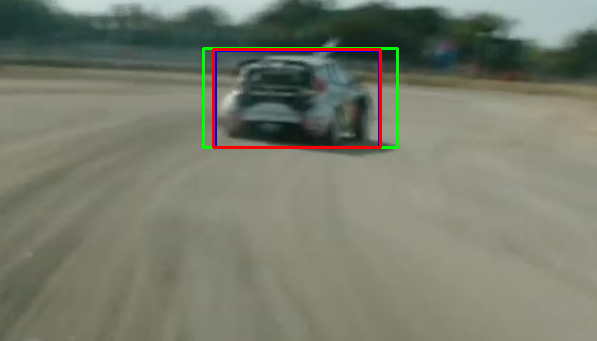}
\includegraphics[height=1.9cm, width=2.8cm]{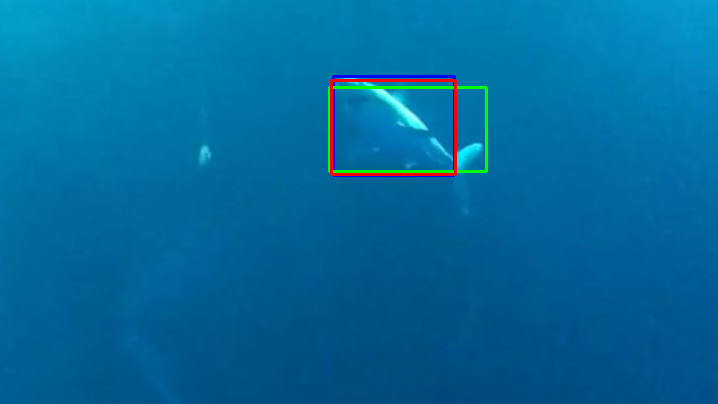}
\includegraphics[height=1.9cm, width=2.8cm]{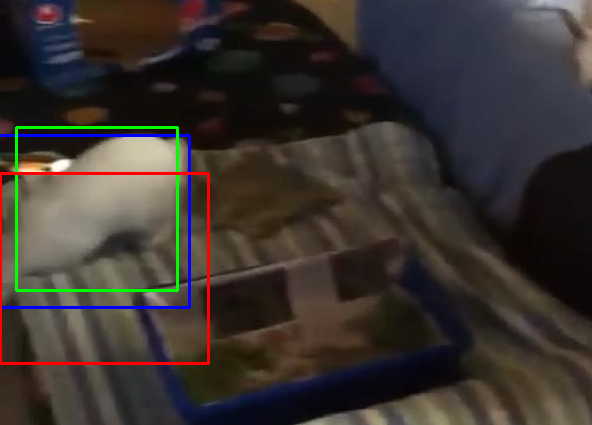}
\caption{Channel dropout}
\label{srpn_ex_channel}
\end{subfigure}
\begin{subfigure}{0.49\textwidth}
\centering
\includegraphics[height=1.9cm, width=2.8cm]{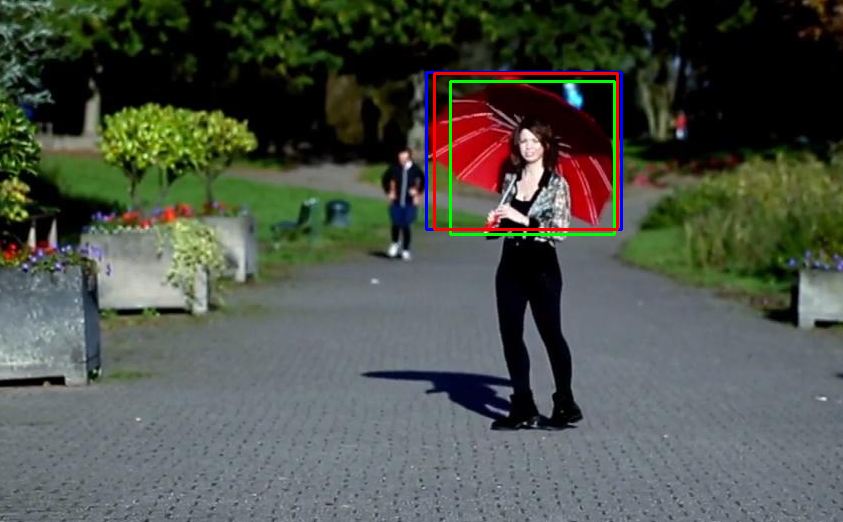}
\includegraphics[height=1.9cm, width=2.8cm]{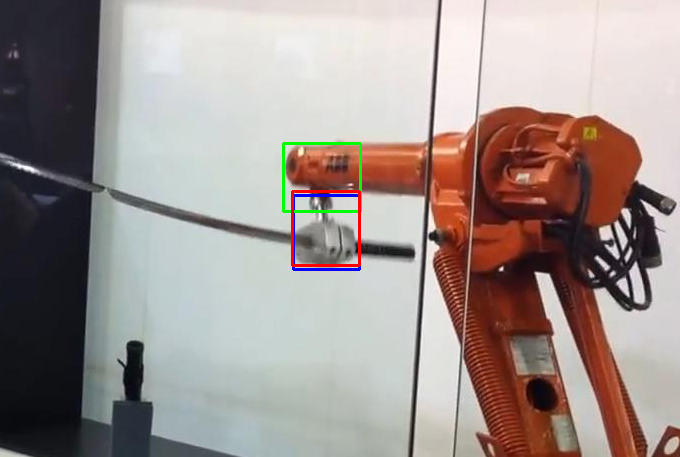}
\includegraphics[height=1.9cm, width=2.8cm]{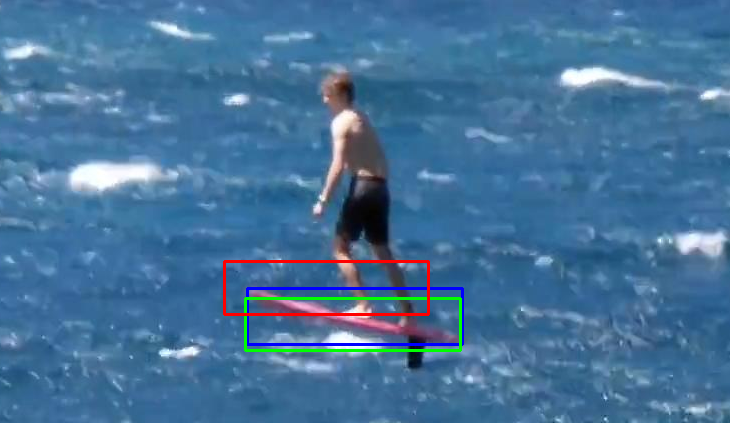}
\caption{Segment dropout}
\label{srpn_ex_segment}
\end{subfigure}
\begin{subfigure}{0.49\textwidth}
\centering
\includegraphics[height=1.9cm, width=2.8cm]{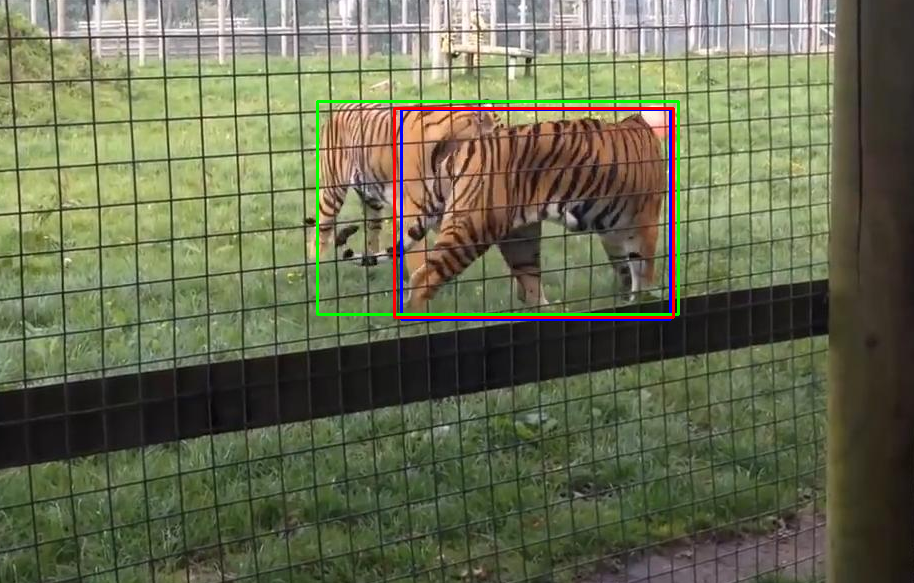}
\includegraphics[height=1.9cm, width=2.8cm]{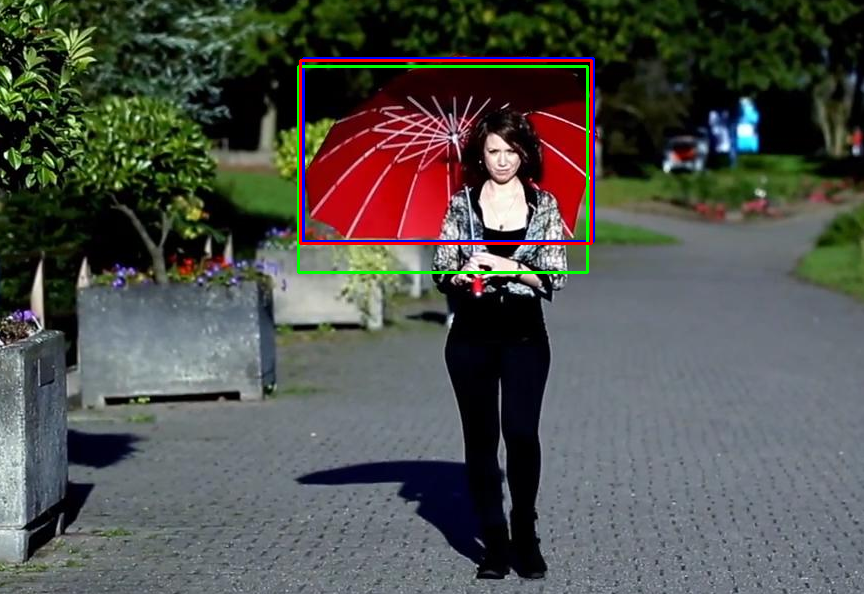}
\includegraphics[height=1.9cm, width=2.8cm]{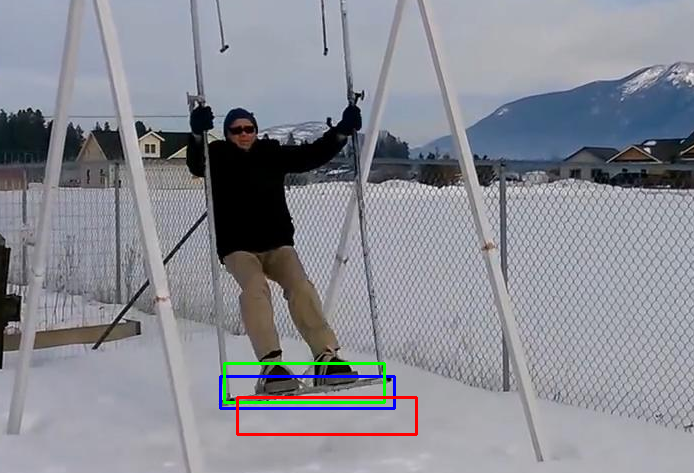}
\caption{Slice dropout}
\label{srpn_ex_slice}
\end{subfigure}
\caption{Predictions obtained using SiamRPN-SD (red), base SiamRPN++ (green) and ground-truth (blue) for frames sampled from LaSOT dataset. For every dropout approach, the two predictions from the left demonstrate improvement obtained using SiamRPN-SD and the rightmost frames show cases where structured dropouts fail.}
\label{fig_srpn_ex}
\end{figure}

\begin{figure}
\centering
\begin{subfigure}{0.24\textwidth}
\centering
\includegraphics[scale=0.17]{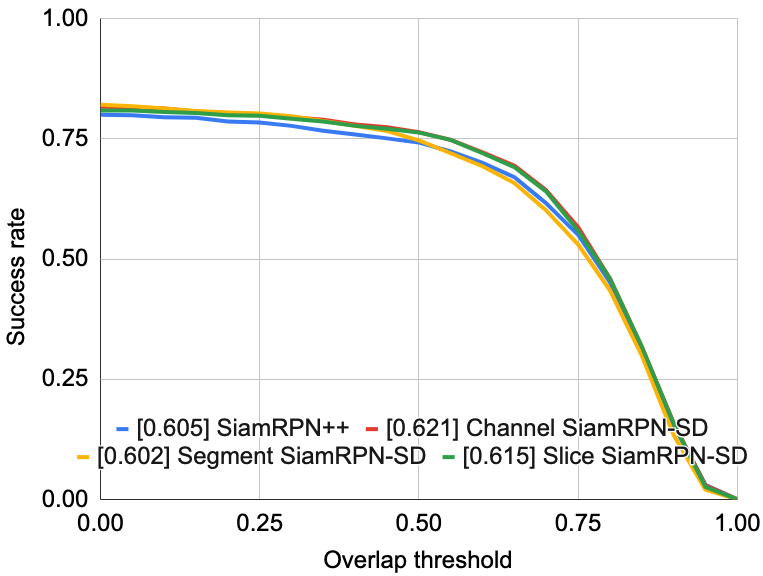}
\end{subfigure}
\begin{subfigure}{0.24\textwidth}
\centering
\includegraphics[scale=0.17]{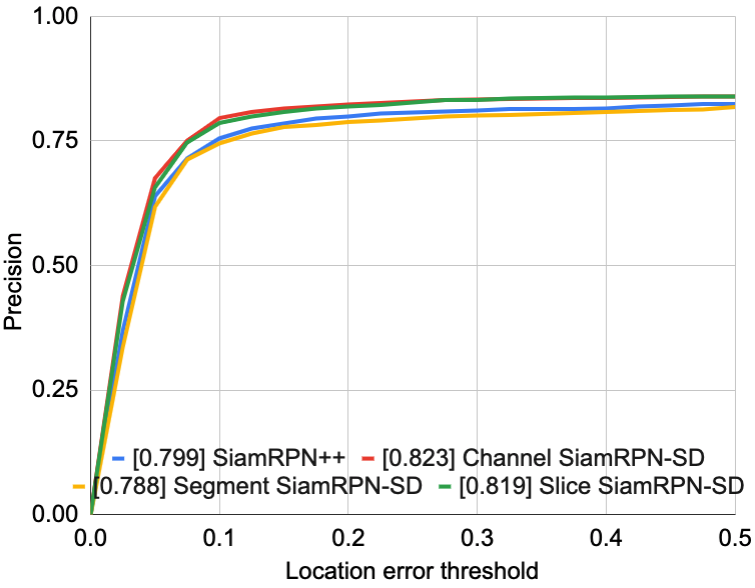}
\end{subfigure}
\caption{Performance plots of OPE on UAV123 obtained using SiamRPN-SD.}
\label{uav_ope_plots}
\end{figure}

\begin{figure}
\centering
\begin{subfigure}{0.24\textwidth}
\centering
\includegraphics[scale=0.18]{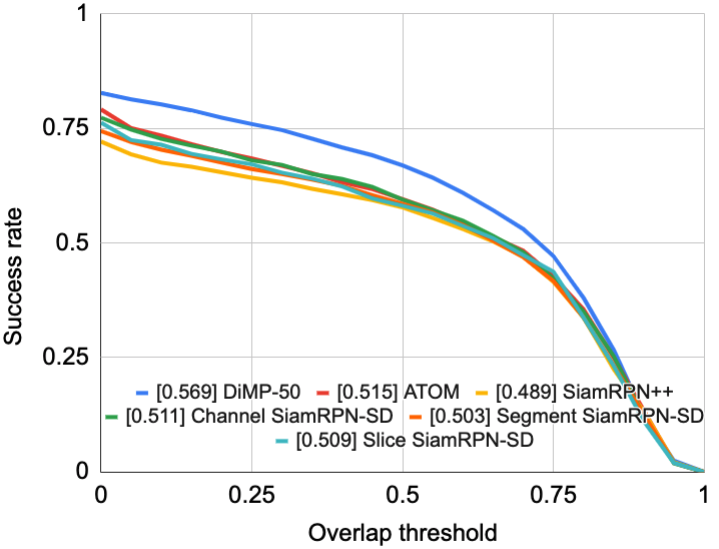}
\end{subfigure}
\begin{subfigure}{0.24\textwidth}
\centering
\includegraphics[scale=0.17]{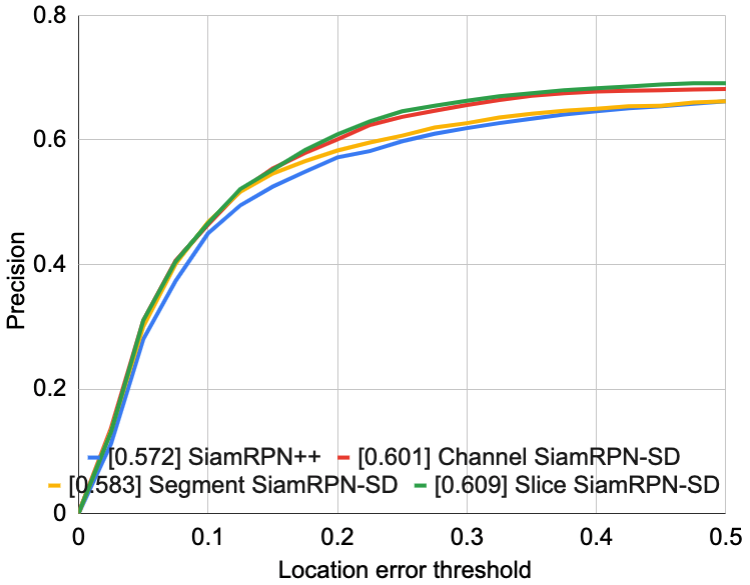}
\end{subfigure}
\caption{Performance plots of OPE on LaSOT obtained using SiamRPN-SD.}
\label{lasot_ope_plots}
\end{figure}

\textbf{SiamRPN++ with Structured Dropouts.} We further perform detailed analysis of structured dropouts with SiamRPN++, and refer it as SiamRPN-SD. Experiments have been performed on OTB2015, UAV123, VOT2018 and LaSOT and details on the results are shown in Table \ref{table_results_siamrpnpp} and Figures \ref{uav_ope_plots} and \ref{lasot_ope_plots}. We also show some example frames for the three dropouts in Figure \ref{fig_srpn_ex} to present a qualitative analysis on performance. 

\emph{Channel SiamRPN-SD. } The channel dropout implementation improves performance on OTB2015 as well as VOT2018. While the improvements on VOT2018 are small, absolute improvements of approximately 2\% are obtained on precision as well as accuracy scores. Similar gains are also obtained on UAV123, as can be seen from the precision and success plots of OPE in Figure \ref{uav_ope_plots}. We see that SiamRPN-SD with channel dropouts performs consistently better than SiamRPN++ across different values of overlap threshold as well as the location error threshold. Similar observations are also made on LaSOT dataset, as shown in Figure \ref{lasot_ope_plots}. While channel dropout helped to improve the prediction in several frames, it degraded the performance in a few others. Examples from LaSOT dataset are shown in Figure \ref{srpn_ex_channel}. 

\emph{Segment SiamRPN-SD. } With segment dropout, we observe improvement in performance on OTB100, especially a 1.5\% improvement on precision score. However, the performance seems to drop slightly on VOT2018 dataset. From Figure \ref{uav_ope_plots}, we observe that the performance values drop on UAV123 compared to the baseline tracker. On LaSOT dataset (Figure \ref{lasot_ope_plots}), improvements of 2\% and 1\% are obtained for the success and precision scores, respectively. To summarize SiamRPN-SD with segment dropouts does not consistently improve performance on all datasets, and can have adverse affects as well. Figure \ref{srpn_ex_segment} shows example frames from LaSOT obtained using segment dropouts, with the rightmost showing a failure case.

\emph{Slice SiamRPN-SD. } SiamRPN-SD with slice dropouts has been found to improve tracker performance on all the 4 datasets. For OTB2015, these are 2.4\% and 1.8\ for success and precision scores, respectively. Similar to channel dropouts, negligible improvements are obtained on VOT0218 dataset. On UAV123, the improvements are 1\% and 2\% on success and precision scores. From Figure \ref{lasot_ope_plots}, it can be seen that slice dropout delivers an increase of 3.5\% on the precision score for LaSOT dataset. Example predictions from LaSOT obtained using slice droput are shown in Figure \ref{srpn_ex_slice}.

Summarizing over the results for SiamRPN-SD, it can be inferred that channel as well as slice dropouts can improve tracker's performance. The inconsistency with segment dropout can be due to the limited number of samples for each pass. Nevertheless, our structured dropouts perform better than the equivalent MC-dropout implementations.

%% file: discuss.tex
\emph{Choice of Structured Dropout. } Based the experiments related to SiamFC-SD and SiamRPN-SD, we argue that both, channel as well as slice dropouts can be used with siamese trackers for improving tracker's performance. Our segment dropouts do not show expected improvements in some cases. Since slice dropouts are a tailored non-stochastic version of segment dropouts, we believe that larger number of dropout samples could help in further improving the segment dropouts.

Among the channel and slice dropouts, there is no one winner. As stated earlier, both are defined for different types of occlusion, and for a large dataset such as LaSOT, where both and many more challenges can occur, possibly a combination of both might perform better. However, this aspect has not been explored yet and will be a part of a future study. It is also important that the outputs of dropouts are combined using an additional encoding in the latent space, rather than explicit sampling as in \cite{Miller2018icra} and the baseline stated in this paper.

\emph{Effect on inference speed. } Since structured dropouts are meant to be applied in the latent space and to be combined using the additional small encoding architecture, these are accompanied by only very small reduction in inference speed, thereby keeping them still within the acceptable regime. For example, SiamFC-SD with channel dropouts and $n = 20$ operated at around 50 fps, while the base siamFC had the speed of 75 fps. On the other hand, for the explicit sampling based dropout as well as the MC-dropout implementation, inference time grows linearly with the number of samples. For the equivalent explicit implementation, the inference speed dropped to around 12 fps.

%% file: conclusion.tex
% \str{We do this later and after everything else is done. But it should not be more than 7-10 lines.}

In this paper, we present structured dropouts, semantically designed dropout methodology to tackle the effect of occlusion in object tracking. Unlike the common forms of dropout, structured dropouts  are intended to mimic occlusion in the latent space. It can be interpreted as a methodological block which can be added to any siamese tracker at a relatively small increase in inference time, while improving the performance of trackers for up to 3\%. Through combining output from multiple dropout passes using an additional encoding network, the model can be trained end-to-end, and we demonstrate that such a strategy can improve the robustness of the model against occlusion. In fact, as occlusion can have unlimited variations in their appearances as any object can occlude another, we argue it is the only known feasible approach to handling occlusion specifically. The performance improvements on several tracking benchmarks presented in this study demonstrate the applicability as well as the robustness of our approach.